\documentclass[letterpaper, 10 pt, conference]{ieeeconf} 
\IEEEoverridecommandlockouts     

\usepackage{graphics} 
\usepackage{graphicx}   
\usepackage{amsmath} 
\usepackage{amssymb}  
\usepackage{mathtools, cuted}
\usepackage{interval}
\usepackage{nicefrac}
\usepackage[noadjust]{cite}
\usepackage{breqn}
\usepackage{color, colortbl}
\usepackage{derivative}
\usepackage{siunitx}
\usepackage{multirow}
\usepackage{hyperref}
\usepackage{cleveref}
\usepackage{mdframed} 

\usepackage{placeins}  

\renewcommand{\vec}[1]{\mathbf{\boldsymbol{#1}}}

\usepackage{marginnote}

\usepackage[dvipsnames]{xcolor}

\title{16 Ways to Gallop: Energetics and Body Dynamics of \\ High-Speed Quadrupedal Gaits}

\author{Yasser G. Alqaham$^*$, Jing Cheng$^*$, and Zhenyu Gan, \textit{ Member, IEEE}
\thanks{All authors are with the Department of Mechanical and Aerospace Engineering, Syracuse University, Syracuse, NY 13244 \texttt{\{ygalqaha, jcheng13, zgan02\}@syr.edu}. }
\thanks{$^*$The authors contribute equally to this paper.}
\thanks{This work was supported by Syracuse University and the NSF Energy Storage Engine in Upstate New York (SP-33598-1).}
}

\begin{document}

\maketitle
\thispagestyle{empty}
\pagestyle{empty}

\begin{abstract}
Galloping is a common high-speed gait in both animals and quadrupedal robots, yet its energetic characteristics remain insufficiently explored. This study systematically analyzes a large number of possible galloping gaits by categorizing them based on the number of flight phases per stride and the phase relationships between the front and rear legs, following Hildebrand’s framework for asymmetrical gaits.  
Using the A1 quadrupedal robot from Unitree, we model galloping dynamics as a hybrid dynamical system and employ trajectory optimization (TO) to minimize the cost of transport (CoT) across a range of speeds. Our results reveal that \textbf{rotary and transverse gallop footfall sequences exhibit no fundamental energetic difference}, despite variations in body yaw and roll motion. However, \textbf{the number of flight phases significantly impacts energy efficiency}: galloping with no flight phases is optimal at lower speeds, whereas galloping with two flight phases minimizes energy consumption at higher speeds.  
We validate these findings using a Quadratic Programming (QP)-based controller, developed in our previous work, in Gazebo simulations. These insights advance the understanding of quadrupedal locomotion energetics and may inform future legged robot designs for adaptive, energy-efficient gait transitions.
\end{abstract}

\section{Introduction}
\label{sec:intro}
\begin{figure}[t!]
\centering
\includegraphics[width=1\columnwidth]{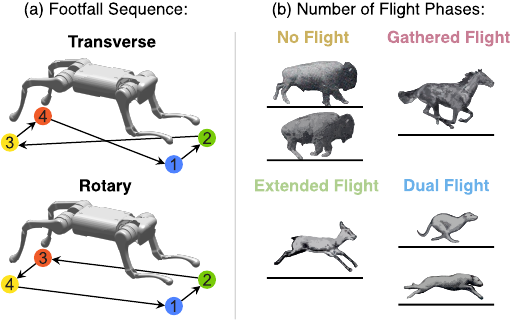}
\caption[]{(a) Footfall sequence variations between transverse and rotary galloping gaits, with foot placement color-coded: right-fore foot (blue), left-fore foot (green), right-hind foot (yellow), and left-hind foot (red). (b) Classification of galloping gaits based on suspension (flight) phases, including no suspension, gathered suspension, extended suspension, and dual suspension~\cite{Hildebrand1977}. Animal images sourced from~\cite{Muybridge1902}.}
\label{fig:gallop_intro}
\vspace{-5mm}
\end{figure}

Quadrupedal locomotion has long fascinated researchers in both biology and robotics due to its complexity and adaptability. Among various gaits, galloping stands out as a high-speed strategy characterized by asymmetrical footfall patterns and aerial phases, enabling rapid and agile movement. Understanding the mechanics and energetics of galloping is crucial for developing quadrupedal robots capable of efficient and dynamic locomotion.
Galloping is prevalent among mammals such as cheetahs, horses, and dogs. This gait follows a distinct footfall sequence, often incorporating one or more aerial phases per stride. As illustrated by Fig.~\ref{fig:gallop_intro}, Hildebrand's seminal work categorized asymmetrical gaits based on limb phase relationships, distinguishing between \textbf{transverse gallops}, where the hind limb contacts are followed by the ipsilateral forelimb, and \textbf{rotary gallops}, which exhibit a circular limb sequence~\cite{Hildebrand1977,Hildebrand1989}. These patterns affect speed and stability; for instance, cheetahs employ rotary gallops for sprinting, while horses often use transverse gallops for endurance running~\cite{Biancardi2012}.
Additionally, the frequency of the flight phases plays a pivotal role in shaping the overall energy expenditure~\cite{McMahon1985,Walter2007}. Galloping with no flight phases is more energy-efficient at lower speeds, while incorporating one or two flight phases enhances efficiency at higher speeds. This adaptability allows animals to optimize gait selection based on environmental and metabolic constraints.

Translating biological galloping into robotics presents significant challenges. Early efforts focused on walking and trotting, but achieving high-speed, agile locomotion required exploring galloping gaits. The MIT Cheetah robots demonstrated stable galloping using model-based control~\cite{DiCarlo2018,Kim2019HighlyDQ}. Another example, the Scout II quadruped, implemented a rotary gallop using passive dynamics and minimal sensory feedback~\cite{Poulakakis2003}, highlighting the role of mechanical design in gait stability.
Recent advances have further improved robotic speed and agility. The Black Panther II, developed by Zhejiang University, achieves 10.4 m/s while sprinting 100 meters in under 10 seconds~\cite{blackpanther2024}. Similarly, the HOUND robot, developed by KAIST, set a Guinness World Record with a 100-meter sprint time of 19.87 seconds~\cite{guinness2023fastest}. These developments demonstrate the growing capabilities of quadrupedal robots. However, both robots use trotting, an intermediate-speed gait that prioritizes stability over maximal velocity. In contrast, animals rely on galloping to reach their highest speeds, emphasizing the need for further research into \textbf{energy-efficient galloping gaits for robotic quadrupeds}.

Replicating an efficient galloping gait remains an ongoing challenge. Maintaining dynamic stability at high speeds requires precise coordination of limb movements and body dynamics, necessitating robust control strategies. Additionally, robotic galloping often incurs a higher \textbf{cost of transport} (CoT) than biological counterparts, highlighting the need for optimization in both control algorithms and mechanical design.
The number of flight phases plays a critical role in energy efficiency. While increased aerial phases enable higher speeds, they also require greater energy input. Optimizing this balance is essential for achieving both speed and efficiency. Furthermore, transitioning between trotting and galloping in unstructured environments introduces additional challenges, as terrain variability affects gait stability.

This work systematically analyzes galloping gaits in quadrupedal robots, modeling galloping dynamics as a hybrid system and employing trajectory optimization to minimize CoT across different speeds. We categorize galloping gaits based on the number of flight phases and investigate their impact on energy efficiency.
Our key contributions include: (i) \textbf{Comprehensive gait analysis}: A detailed examination of transverse and rotary galloping sequences and their energetic profiles. (ii) \textbf{Optimization framework}: Implementation of trajectory optimization to identify gait parameters that minimize energy consumption, informing efficient locomotion strategies for robots.
This study bridges the gap between biological insights and robotic applications, advancing the development of quadrupedal robots capable of high-speed, energy-efficient locomotion.

\section{Methods}
This section outlines the methodology adopted in this study. First, we describe the different types of galloping gaits considered. Next, we present the modeling of the quadrupedal robot A1, followed by an overview of its galloping hybrid dynamics. We then detail the formulation of the nonlinear trajectory optimization problem.

\label{sec:methods}

\subsection{Footfall Sequence of Galloping Gaits}
\vspace*{-0.2in}
\begin{figure}[ht!]
\centering
\includegraphics[width=0.95\columnwidth]{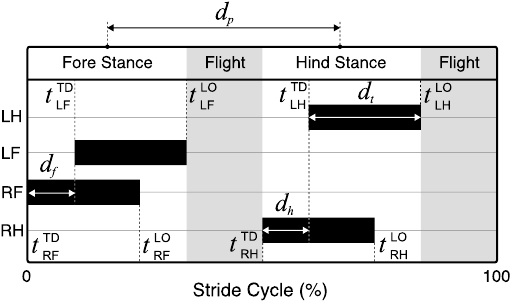}
    \caption{Gait diagrams of representative galloping gaits over one normalized stride, from 0\% (beginning of the stride, defined by right front leg touchdown) to 100\% (end of the stride).}
    \label{fig:gait_diagram}
\vspace*{-0.1in}
\end{figure}

The footfall pattern of a gait is uniquely defined by its footfall sequence, represented as $\vec{t} = [t^{\text{TD}}_{\scriptscriptstyle\text{LH}}, t^{\text{LO}}_{\scriptscriptstyle\text{LH}}, t^{\text{TD}}_{\scriptscriptstyle\text{LF}}, t^{\text{LO}}_{\scriptscriptstyle\text{LF}}, t^{\text{TD}}_{\scriptscriptstyle\text{RF}}, t^{\text{LO}}_{\scriptscriptstyle\text{RF}}, t^{\text{TD}}_{\scriptscriptstyle\text{RH}}, t^{\text{LO}}_{\scriptscriptstyle\text{RH}}]^T$. Each timing value $t^j_i$ (where $i \in \{\text{LH}, \text{LF}, \text{RF}, \text{RH}\}$ and $j \in \{\text{TD}, \text{LO}\}$) is normalized by the stride duration $t_{\text{stride}}$ and ranges from 0 to 100\%. The initials L, R, F, and H represent left, right, fore, and hind feet, respectively, while TD and LO denote touchdown and lift-off events. Using Hildebrand's convention~\cite{Hildebrand1977}, footfall sequences are illustrated with black bars for ground contact and white spaces for suspension, with total stride duration spanning the full sequence (e.g., Fig.~\ref{fig:gait_diagram}). To quantify different galloping gaits, four phase-based metrics are introduced:
\begin{figure*}[htbp]
\centering
\includegraphics[width=1.0\textwidth]{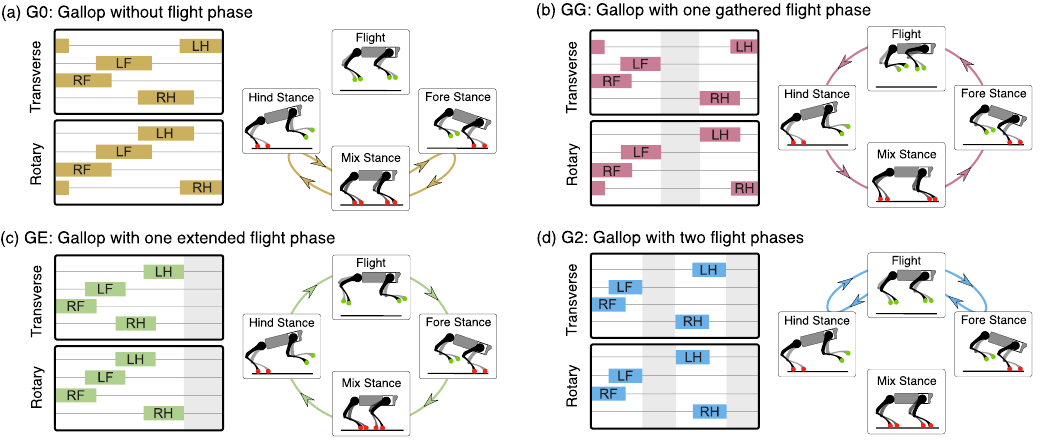}
\caption[Galloping Gaits]{Illustration of different variations of galloping gaits. Panels (a)–(d) depict gaits with varying numbers of flight phases (shaded in gray). Each gait is shown under two primary gallop styles: transverse (top) and rotary (down). The subplot on the right represents the stride cycle in terms of hybrid dynamics and phase transitions. Theoretically, our analysis identifies \textbf{16 possible galloping footfall sequences}. However, this study focuses exclusively on right-fore (FR) leading gaits due to energetic symmetries~\cite{Gan2018, ding2023breaking,Apraez2025}.}
\label{fig:Galloping_Gaits}
\vspace{-6mm}
\end{figure*}

\begin{itemize}
    \item \textbf{Duty factor:} $d_t \in [0, 1) := (t^{\text{LO}}_i - t^{\text{TD}}_i) \mod 1$. When \( t^{\text{LO}}_i < t^{\text{TD}}_i \), the modulo operation accounts for the periodicity of the stride cycle by effectively adding 1 before taking the difference, ensuring a consistent and normalized duty factor between 0 and 1, representing the relative stance duration of a leg, assuming all four legs have similar duty factors~\cite{Hildebrand1977}.
    
    \item \textbf{Foreleg phase lag:} $d_f \in [-0.5, 0.5) := t^{\text{TD}}_{\scriptscriptstyle\text{LF}} -  t^{\text{TD}}_{\scriptscriptstyle\text{RF}} $ \footnote{Similar modulo operation is used for all three phase lags $d_f$, $d_h$, and $d_p$, for example: 
    \[
    d_f := \left( (t^{\text{TD}}_{\scriptscriptstyle\text{LF}} - t^{\text{TD}}_{\scriptscriptstyle\text{RF}} + 0.5) \mod 1 \right) - 0.5
    \]
    }.
    
    \item \textbf{Hindleg phase lag:} $d_h \in [-0.5, 0.5) :=t^{\text{TD}}_{\scriptscriptstyle\text{LH}} -  t^{\text{TD}}_{\scriptscriptstyle\text{RH}} $.
    
    \item \textbf{Fore-hind phase lag:} $d_p \in [-0.5, 0.5) := \nicefrac{[(t^{\text{LO}}_{\scriptscriptstyle\text{LH}} - t^{\text{TD}}_{\scriptscriptstyle\text{LH}}) + (t^{\text{LO}}_{\scriptscriptstyle\text{RH}} - t^{\text{TD}}_{\scriptscriptstyle\text{RH}}) - 
    (t^{\text{LO}}_{\scriptscriptstyle\text{LF}} - t^{\text{TD}}_{\scriptscriptstyle\text{LF}}) - (t^{\text{LO}}_{\scriptscriptstyle\text{RF}} -t^{\text{TD}}_{\scriptscriptstyle\text{RF}})]}{2}$, indicating the phase lag between midstance of the fore and hind pairs.
\end{itemize}

With the metrics defined above, galloping gaits in prior studies \cite{Hildebrand1977, Park2015, Herr2001, Poulakakis2003} can be categorized using three criteria:

\subsubsection{Gallop Footfall Type} If $d_f$ and $d_h$ have the same sign, it is classified as a \textbf{Transverse gallop}; if they have opposite signs, it is a \textbf{Rotary gallop}. Note that a transverse gallop can be transformed into a rotary gallop, and vice versa, by reversing the motion of \textbf{a single leg pair}. However, this modification alters the torso's oscillation modes, as detailed in Section~\ref{sec:body_motion}.

\subsubsection{Number of Flight Phases} Based on suspension phases \cite{ding2023breaking, Alqaham2024}, galloping gaits are categorized as follows:

\paragraph{G0} $:= \{(d_t, d_p, d_f, d_h) \,|\, 1-d_t < d_p < d_t, \, d_f \neq 0, \, d_h \neq 0\}$. This gallop has two mixed stance phases and \textbf{no flight phases} as shown in Fig.~\ref{fig:Galloping_Gaits}(a).
        
\paragraph{GG} $:= \{(d_t, d_p, d_f, d_h) \,|\, d_p < d_t, d_p < 1-d_t, \, d_f \neq 0, \, d_h \neq 0\}$. As indicated in Fig.~\ref{fig:Galloping_Gaits}(b), this gait features a single flight phase. During aerial suspension, the legs are gathered under the body, hence a \textbf{Galloping gait with a Gathered flight phase}.
      
\paragraph{GE} $:= \{(d_t, d_p, d_f, d_h) \,|\, d_t < d_p, 1-d_t < d_p, \, d_f \neq 0, \, d_h \neq 0\}$. As shown in Fig.~\ref{fig:Galloping_Gaits}(c), this gait also has a single flight phase. However, during suspension, the legs extend outward from the body, resulting in a \textbf{Galloping gait with an Extended flight phase}.
        
\paragraph{G2} $:= \{(d_t, d_p, d_f, d_h) \,|\, d_t < d_p < 1-d_t, \, d_f \neq 0, \, d_h \neq 0\}$. This gait has two flight phases, as illustrated in Fig.~\ref{fig:Galloping_Gaits}(d). Depending on the execution, the torso reaches its highest position either when the legs are gathered or extended, leading to a \textbf{Galloping gait with 2 flight phases}.

\subsubsection{Leading vs. Trailing Foot} The first foot to strike the ground in a couplet is termed the \emph{trailing foot}, while the second is the \emph{leading foot} \cite{Hildebrand1977}. Each galloping gait has two variations depending on which front leg leads. A \textbf{right-fore (FR) leading gait} can be transformed into a left-fore (LF) leading gait, and vice versa, by reversing the motion of \textbf{both leg pairs}.

Theoretically, our analysis identifies 16 possible galloping footfall sequences, determined by three key factors: (i) two primary gallop styles (transverse and rotary), (ii) four distinct classifications based on the number of flight phases, and (iii) two possible leading limb configurations (left-fore (FL) or right-fore (FR)). This combinatorial structure results in a total of $2 \times 4 \times 2 = 16$ distinct sequences.  
However, in this study, we focus exclusively on \textbf{right-fore (FR) leading gaits}. This selection is justified by the presence of energetic symmetries between left- and right-leading gaits, meaning that the energy consumption and dynamic behavior of FL-leading gaits mirror those of their FR-leading counterparts~\cite{Gan2018, ding2023breaking, Apraez2025}. As a result, analyzing only FR-leading gaits provides a complete representation of all unique energetic characteristics while reducing computational redundancy.

\vspace{-1mm}
\subsection{Robot Model}
\label{sec:Robot Model}

The A1 quadrupedal robot, shown in Fig.~\ref{fig:gallop_intro}(a), consists of a main body with four articulated limbs, each featuring three actuated joints driven by electric motors. Each limb includes hip abduction/adduction, hip flexion/extension, and knee flexion/extension joints. The system's configuration is represented by the generalized state vector:
\begin{align}
\mathbf{q} = \begin{bmatrix} \mathbf{r}_{b} \\ \boldsymbol{\theta}_{b} \\ \mathbf{q}_{l} \end{bmatrix} \in \mathcal{Q},
\end{align}
where \(\mathbf{r}_{b} \in \mathbb{R}^3\) represents the spatial position of the robot's torso, \(\boldsymbol{\theta}_{b} \in \mathbb{SO}(3)\) denotes its orientation, and \(\mathbf{q}_{l} \in \mathbb{S}^{12}\) corresponds to the limb joint angles.
The system dynamics are governed by the Euler-Lagrange equations:

\begin{align}
\mathbf{M}(\mathbf{q}) \ddot{\mathbf{q}} + \mathbf{C}(\mathbf{q}, \dot{\mathbf{q}}) + \mathbf{G}(\mathbf{q}) = \mathbf{U} + \sum_{i} \mathbf{J}_{i}^{T}(\mathbf{q}) \boldsymbol{\lambda}_{i},
\label{eq:EoM}
\end{align}
where: \(\mathbf{M}(\mathbf{q}) \in \mathbb{R}^{18 \times 18}\) is the mass-inertia matrix, \(\mathbf{C}(\mathbf{q}, \dot{\mathbf{q}}) \in \mathbb{R}^{18}\) represents Coriolis and centrifugal effects, \(\mathbf{G}(\mathbf{q}) \in \mathbb{R}^{18}\) accounts for gravitational forces, \(\mathbf{U} \in \mathbb{R}^{18}\) is the extended actuator torque vector, with the first six elements set to zero and the last twelve elements corresponding to the actual actuation torques, \(\sum_{i} \mathbf{J}_{i}^{T}(\mathbf{q}) \boldsymbol{\lambda}_{i}\) represents the summation over all external forces projected through their respective contact Jacobians.

\subsection{Hybrid Dynamics}

Galloping in quadrupedal locomotion follows a hybrid dynamical structure, where continuous motion phases are intermittently interrupted by discrete transitions dictated by foot-ground interactions. Specifically, each leg alternates between \textit{stance}, where the foot maintains ground contact, and \textit{swing}, when the leg is airborne. These transitions introduce velocity discontinuities, necessitating a mathematical framework that integrates both continuous dynamics and impact-driven state changes, as in \cite{Li2020,Singh2022,Alqaham2024}.

When the $i$-th leg is in stance, a holonomic constraint $\vec{g}_i(\vec{q}) = p_i$, where $\vec{g}_i(\vec{q})$ represents the foot’s forward kinematics and \( p_i \) denotes its contact position, is enforced to ensure that the foot remains stationary. Additionally, ground reaction forces must be nonnegative to maintain realistic contact dynamics, ensuring that the foot only exerts compressive forces on the ground. The resulting ground reaction force $\boldsymbol{\lambda}$ governs the system’s motion until the foot lifts off, at which point its associated holonomic constraint is removed.

During swing, a unilateral constraint $\vec{g}_i(\vec{q}) > 0$ is applied to prevent ground penetration, ensuring that the foot remains above the surface until touchdown. At touchdown, impact dynamics cause an instantaneous velocity change, generating an impulse force $\Lambda$ that modifies the system’s velocities as the foot abruptly comes to rest. This impact event follows an impulse-momentum relationship, ensuring consistency between pre- and post-impact states:
\begin{align}
\dot{\vec{q}}^+ = \vec{G}(\vec{q}^-, \dot{\vec{q}}^-),
\label{eq:ResetMap}
\end{align}
where $\dot{\vec{q}}^+$ represents the system’s velocity immediately after impact, while $\vec{q}^-$ and $\dot{\vec{q}}^-$ denote the configuration and velocity just before contact.

\subsection{Trajectory Optimization}

Our objective is to determine optimal periodic galloping trajectories across a range of speeds. To achieve this, we employ a hybrid trajectory optimization approach, which integrates multiple trajectory optimization formulations, each constrained by its respective dynamics and contact conditions. A widely used method in this context is direct collocation, which transforms the hybrid optimization problem into a constrained nonlinear program (NLP).
We formulate the discrete-time NLP optimization problem as:

\begin{subequations}
\begin{align}
    & \underset{\vec{\gamma}}{\operatorname{\text{argmin}}}
    && \sum_{k=1}^{N}{g_{k}(\vec{\gamma})}, \\
    & \text{subject to}
    && (\vec{q}_{k+1},\dot{\vec{q}}_{k+1}) = f(\vec{q}_{k},\dot{\vec{q}}_{k},\vec{u}_{k}), \\
    &&& \vec{\gamma}_{k} \in \vec{\Gamma}. 
\end{align}
\end{subequations}

where $\vec{\gamma}$ represents the design variables at each mesh point, $g_k(\vec{\gamma})$ is the objective function, $(\vec{q}_{k+1},\dot{\vec{q}}_{k+1}) = f(\vec{q}_{k},\dot{\vec{q}}_{k},\vec{u}_{k})$ defines the discretized system dynamics, and $\vec{\Gamma}$ is the feasible design space, incorporating equality and inequality constraints.

\subsubsection{Objective Function}
The optimization aims to achieve the desired speeds while minimizing the work-based cost of transport (CoT), accounting for both positive and negative mechanical work:

\begin{equation}
\label{eq:cost function}
    \text{CoT} =\frac{\int_{t_o}^{t_f} | \vec{u} \cdot \vec{\dot{q}}_{l}| \:d t}{mg \left[ \: {x} {\scriptstyle (t_f)} - {x}{\scriptstyle (t_o)} \: \right]}  
\end{equation}

where $t_o$ and $t_f$ denote the beginning and end of a full stride cycle. The vectors $\vec{u}$ and $\vec{\dot{q}}_{l}$ represent the joint torques and velocities, respectively. The denominator consists of the robot’s weight, $mg$, and the horizontal displacement of the torso during the stride, ${x} {\scriptstyle (t_f)} - {x}{\scriptstyle (t_o)}$.

\subsubsection{Constraints}
The optimization process enforces several constraints to ensure feasibility while satisfying motion requirements and physical limitations:

\vspace{2mm}
\textbf{Optimization Constraints:}
\begin{itemize}
    \item \textbf{Dynamics constraints:} Enforced as per Eq.~(\ref{eq:EoM}).
    \item \textbf{Holonomic constraints in stance:} $\vec{g}_i(\vec{q}) = p_i$ ensures foot placement.
    \item \textbf{Unilateral constraints in swing:} $\vec{g}_i(\vec{q}) > 0$ prevents ground penetration.
    \item \textbf{Average sagittal speed:}  
    $\bar{\dot{x}} = \frac{{x}(t_f) - {x}(t_0)}{t_f - t_0}$ m/s.
    \item \textbf{Periodicity constraints:}
    \begin{itemize}
        \item[--] $\mathbf{q}(t_0) = \mathbf{q}(t_f)$ (cyclic motion),
        \item[--] ${x}(t_0) \ne {x}(t_f)$ (ensures forward progression).
    \end{itemize}
    \item \textbf{Friction cone:}  
    $||{\lambda}_{i,x} + {\lambda}_{i,y}||_2 - \mu {\lambda}_{i,z} \leq 0$.
    \item \textbf{Swing foot velocity at touchdown:} Ensures controlled impact at foot-strike.
    \item \textbf{State, input, and phase duration limits:}
    \begin{itemize}
        \item[--] $\mathbf{q}_{\text{min}}\leq\mathbf{q}\leq\mathbf{q}_{\text{max}}$ (joint limits),
        \item[--] $|\dot{\mathbf{q}}|\leq\dot{\mathbf{q}}_{\text{lim}}$ (velocity limits),
        \item[--] $|\vec{u}|\leq\vec{u}_{\text{lim}}$ (torque limits),
        \item[--] $\mathbf{t}_{\text{min}}\leq\mathbf{T}\leq\mathbf{t}_{\text{max}}$ (phase duration limits).
    \end{itemize}
    \item \textbf{Initial body yaw:} Enforced as $\gamma = 0$ to standardize trajectory orientation.
\end{itemize}

\vspace{2mm}

\begin{figure}[t!]
\centering
\includegraphics[width=1\columnwidth]{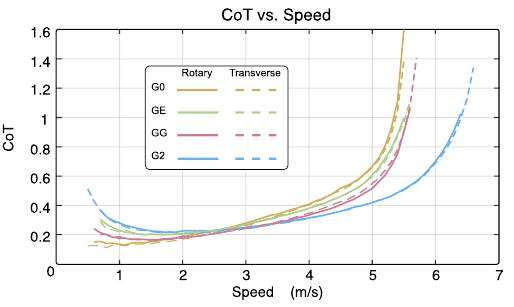}
\caption[CoT]{CoT as a function of speed for different galloping gaits, comparing rotary (solid lines) and transverse (dashed lines) patterns. G0 is shown in yellow, GE in green, GG in red, and G2 in blue. As speed increases, CoT rises across all gaits, with G0 exhibiting the highest CoT at higher speeds, followed by GG and GE, while G2 maintains the lowest CoT. Minimal differences exist between rotary and transverse variations, indicating similar energy efficiency trends across all gaits.}
\label{fig:GCoT}
\vspace{-5mm}
\end{figure}

Here, $\lambda_{i}$ represents the ground reaction force of the $i$-th contact foot. The NLP is implemented using the open-source FROST framework~\cite{hereid2017frost} and solved via Ipopt~\cite{Wchter2005}. The full implementation, including the TO source code and documentation, is available in our GitHub repository \footnote{\url{https://github.com/DLARlab/16WaystoGallop.git}}.

\begin{figure*}[htbp]
\centering
\includegraphics[width=2\columnwidth]{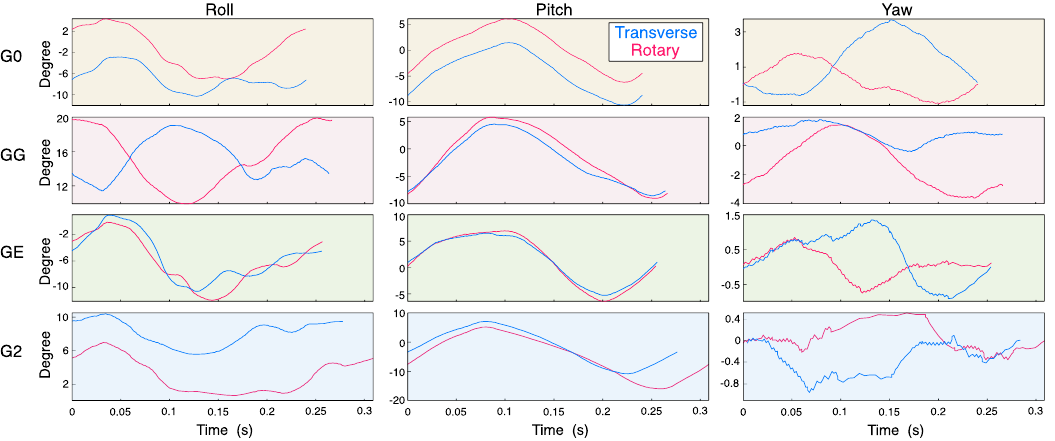}
\caption[Optimal Gaits]{Torso roll (left), pitch (middle), and yaw (right) angles for different galloping gaits (G0, GG, GE, and G2). Blue and magenta curves represent transverse and rotary footfall patterns, respectively. Roll motion exhibits significant variation between footfall patterns, particularly in GG, while pitch motion remains nearly identical across all gaits.}
\label{fig:body_rotations}
\vspace{-5mm}
\end{figure*}

\section{Results}
\label{sec:results}

This section analyzes the CoT across all galloping gaits over a range of speeds, examines the torso's rotational motion, and presents simulation results for G2 galloping in rotary and transverse patterns.

\subsection{Cost of Transport}

Fig.~\ref{fig:GCoT} illustrates how CoT varies with forward speed for all galloping gaits presented in this study. The analyzed speed range extends from 0.5 m/s to 6.6 m/s in increments of 0.1 m/s. The optimization process searches for feasible gaits between 0 and 8 m/s, initializing at 2 m/s and expanding to neighboring speeds to ensure smooth transitions. Gait feasibility is constrained by physical limitations at both slow and high speeds—prolonged stance phases at low speeds push joint limits, while high-speed motion is restricted by hip and knee torque saturation. Each optimization required over 48 hours on an Intel\textregistered\ Core\textsuperscript{TM} i7-10700F @ 2.9 GHz desktop computer.

The feasible speed range for each gait differs. G0 is viable between 0.5 m/s and 5.5 m/s, while GG and GE operate from 0.6 m/s to 5.8 m/s. G2 has the broadest range, from 0.5 m/s to 6.6 m/s.  
A comparison of transverse and rotary galloping reveals nearly identical energy trends, with overlapping CoT curves across all speeds, indicating minimal differences in energy efficiency. This suggests that the choice between transverse and rotary gaits may be driven more by stability, terrain conditions, and biomechanical considerations than by energy consumption. Consequently, we do not differentiate between them in subsequent discussions.

\subsubsection{At speeds below 1.5 m/s} \textbf{G0 exhibits the lowest CoT}, making it the most energy-efficient gait. In contrast, G2 is the least efficient in this range, with a CoT of 0.5 at 0.5 m/s compared to 0.18 for G0. GG and GE fall between these extremes, with GG being slightly more efficient than GE.  

\subsubsection{Between 2 m/s and 3 m/s} all gaits converge to a relatively stable CoT of around 0.2, marking an optimal speed range where multiple gait choices yield similar energy costs. However, at 3 m/s, the efficiency trend reverses—G2 becomes more efficient, and all solutions gradually diverge from G0.  

\subsubsection{At speeds above 3 m/s} CoT increases sharply for G0, GG, and GE, though at different rates. G0 and GE experience the steepest rise, making them less efficient beyond this threshold. G2, however, maintains a lower CoT up to 5.5 m/s, where its CoT remains around 0.5, while G0, GG, and GE reach significantly higher values (1.6 and 1, respectively),\textbf{ establishing G2 as the most energy-efficient gait for high-speed locomotion}.  

Beyond 5 m/s, G0 becomes increasingly impractical due to its steep rise in CoT. GG and GE exhibit similar energy trends to G0 with no significant differences. Furthermore, the near-identical energy performance of transverse and rotary galloping suggests no inherent advantage of one over the other in terms of efficiency.

\subsection{Torso's Rotational Motion}
\label{sec:body_motion}

At high speeds, galloping gaits amplify torso rotational motions. Fig.~\ref{fig:body_rotations} illustrates these motions at 5 m/s, comparing roll (left column), pitch (middle column), and yaw (right column) angles across all gaits.

\subsubsection{Roll Motion}  
A clear distinction exists between the rotary and transverse footfall patterns, with roll angles exhibiting phase shifts and amplitude differences. The most pronounced variation occurs in GG (middle row), where the rotary and transverse curves not only differ in amplitude but also follow distinct oscillatory patterns. In contrast, G2, GE, and G0 display smaller variations, though the curves remain misaligned.
Quantitatively, G2 exhibits a roll range of 5.5$^{\circ}$ (1$^{\circ}$ to 6.5$^{\circ}$) for rotary and 4$^{\circ}$ (6$^{\circ}$ to 10$^{\circ}$) for transverse. GG has the largest variation, with rotary roll motion spanning 10$^{\circ}$ (10$^{\circ}$ to 20$^{\circ}$) and transverse fluctuating by 6$^{\circ}$ (-7$^{\circ}$ to -1$^{\circ}$). GE maintains a more balanced roll motion across footfall patterns, both exhibiting 11$^{\circ}$ of variation. G0 follows a similar trend, with 10$^{\circ}$ for rotary and 8$^{\circ}$ for transverse. These results indicate that GG, GE, and G0 induce the most pronounced roll motion, while G2 maintains a more stable roll profile. Notably, in GE, the rotary and transverse footfall patterns produce nearly identical torso roll motion.  
\textbf{Roll motion is strongly influenced by the footfall pattern, particularly in GG, where the rotary and transverse variations exhibit distinct oscillatory behaviors.}  

\subsubsection{Pitch Motion}  
Unlike roll, the pitch angles remain nearly identical between rotary and transverse footfall patterns across all gaits, with overlapping curves indicating minimal variation. This suggests that pitch motion is dictated by the overall galloping gait structure rather than the specific footfall sequence.
In G2, the pitch range is approximately 20$^{\circ}$ (-15$^{\circ}$ to 5$^{\circ}$) for rotary and 16$^{\circ}$ for transverse. GG exhibits a slightly smaller range of 14$^{\circ}$ (-8$^{\circ}$ to 6$^{\circ}$), while GE follows a similar pattern with 12$^{\circ}$ (-6$^{\circ}$ to 6$^{\circ}$) for rotary and 11$^{\circ}$ for transverse. In G0, both footfall sequences yield a pitch range of 10$^{\circ}$, reinforcing the trend that pitch motion is largely unaffected by footfall pattern.  
\textbf{We observe that pitch motion remains consistent across all gaits, with minimal differences between rotary and transverse footfall patterns.}  

\subsubsection{Yaw Motion}  
Yaw oscillations are significantly smaller than roll and pitch motions across all gaits. G2 and GE exhibit minimal yaw variation, fluctuating within -0.8$^{\circ}$ to 0.4$^{\circ}$ (G2) and -0.6$^{\circ}$ to 1.4$^{\circ}$ (GE), making yaw motion nearly negligible for these gaits.
In GG, yaw motion becomes more structured, with a 5$^{\circ}$ range for rotary galloping and 2$^{\circ}$ for transverse, making it the most pronounced among all gaits. G0 follows closely, exhibiting yaw oscillations of 2$^{\circ}$ to 3$^{\circ}$. Notably, the rotary and transverse yaw curves alternate consistently, indicating a phase shift between footfall patterns. However, compared to roll and pitch, yaw motion plays a minor role in overall torso dynamics, particularly in G2, where it is virtually absent.  
\textbf{In short, yaw motion is insignificant compared to roll and pitch, especially in G2, where it is nearly negligible.}


\subsection{Validating Galloping Gaits in Simulation}

\begin{figure}[tbp]
\centering
\includegraphics[width=1\columnwidth]{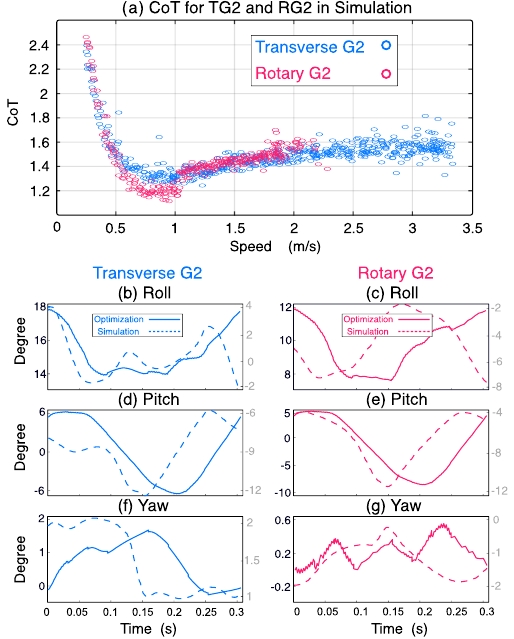}
\caption[G2]{Simulation results for the rotary and transverse G2 galloping gaits. Subplot (a) presents the stride-by-stride CoT, while subplots (b)-(g) illustrate the torso’s angular motion for both optimization results (solid lines, black scale) and simulation results (dashed lines, grey scale).}
\label{fig:6}
\vspace{-5mm}
\end{figure}

To validate our optimal solutions, we conducted simulation tests of the rotary and transverse G2 gaits using the A1 robot in the Gazebo environment, employing a Quadratic Programming (QP)-based locomotion controller developed in our prior work~\cite{Cheng2024}. For each gait, the robot started from rest and gradually accelerated until instability led to a fall at higher speeds. During this process, the CoT was computed stride by stride, as defined in~\eqref{eq:cost function}, and plotted in Fig.~\ref{fig:6}(a).  
As shown in the animations in the accompanying multimedia file, the robot successfully executed galloping gaits with the desired footfall sequences. The CoT trends closely followed those of the optimal solutions in Fig.~\ref{fig:GCoT}, with rotary and transverse patterns exhibiting nearly identical energy consumption. However, the simulated stride-by-stride CoT was significantly higher than the optimized values. Upon inspection, we found that this discrepancy primarily arose from the behavior of the control system in the simulation environment. Specifically, the joint feedback controller and foot placement controllers~\cite{Cheng2024} responded to frequent perturbations with high-frequency corrections. This led to significant joint torque fluctuations, particularly during flight phases, contributing to the observed increase in CoT.

We further compared simulated body rotations with optimization results at approximately 1 m/s. As shown in Fig.~\ref{fig:6}(b)\&(c), roll motion exhibited noticeable oscillations in both transverse and rotary galloping. The simulation and optimization results showed a comparable range of motion, with minimal discrepancies, indicating strong agreement. Pitch motion (Fig.~\ref{fig:6}(d)\&(e)) followed a similar oscillatory pattern across both footfall types and datasets, confirming that pitching motion is unaffected by galloping footfall patterns. However, the range of motion in the simulations was smaller, averaging 7$^{\circ}$ compared to 12$^{\circ}$ in the optimization results.  
Yaw motion (Fig.~\ref{fig:6}(f)\&(g)) exhibited the smallest magnitude compared to roll and pitch, aligning with optimization results. Both footfall patterns showed slight deviations between optimization and simulation, but overall motion trends remained consistent. These results confirm the validity of our optimal gait solutions in simulation.

\section{Conclusions}
\label{sec:conclusions}
This study analyzed the energetic efficiency and body rotational dynamics of various galloping gaits across a range of speeds. Through trajectory optimization and hybrid dynamical modeling, we systematically evaluated the CoT and torso motion in rotary and transverse galloping patterns.

The results indicate that the efficiency of the galloping gaits is highly dependent on speed. At low speeds, \textbf{G0 exhibits the lowest CoT, making it the most energy-efficient option, whereas at higher speeds, G2 becomes the most efficient, maintaining a lower CoT beyond 3 m/s}. This shift may be attributed to the continuous ground contact of G0 at low speeds, which reduces impact forces. In contrast, at higher speeds, galloping gaits like G2, with more frequent flight phases, enable longer strides and leverage ballistic motion, contributing to improved efficiency. Minimal differences were observed between transverse and rotary footfall sequences, suggesting that energy efficiency alone does not favor one pattern over the other.
In terms of body motion, roll dynamics are strongly influenced by footfall patterns, particularly in GG, where rotary and transverse gaits exhibit distinct oscillatory behaviors. Large roll variations in GG, GE, and G0 may challenge lateral stability, requiring active control strategies. In contrast, G2 maintains a more stable roll profile, aiding balance in high-speed locomotion.
Pitch motion remains consistent across all gaits, indicating it is dictated by overall gait structure rather than footfall sequence. This suggests pitch stabilization can be managed uniformly, simplifying sagittal plane control.
Yaw motion is relatively small and negligible in G2, suggesting a minor role in torso dynamics. The reduced yaw oscillations in G2 and GE may enhance heading stability, while GG and G0 exhibit slightly larger yaw deviations, requiring additional control effort.
While \textbf{energy efficiency is speed-dependent, stability considerations favor specific gaits for different tasks}. G2 appears optimal for \textbf{stable, high-speed locomotion}, whereas gaits with greater roll and yaw variations may need enhanced stabilization strategies for reliable movement in unstructured environments. 

To validate these findings, we conducted simulation experiments, which confirmed the results obtained from trajectory optimization. The simulated CoT and torso dynamics closely aligned with the optimization predictions, reinforcing the reliability of our modeling approach. As part of future work, we plan to extend this study by implementing these different galloping gaits on hardware, allowing for real-world validation and further insights into control strategies for quadrupedal robots.


\bibliographystyle{IEEEtran}
\bibliography{References}

\end{document}